\documentclass[10pt]{article}
\usepackage[T1]{fontenc}
\usepackage[utf8]{inputenc}
\usepackage{authblk}
\usepackage{arxiv}

\newcommand{\bs}[1]{\boldsymbol{#1}}
\newcommand{\e}[1]{\cdot 10^{#1}}
\newcommand{\dpar}[2]{\frac{\partial #1}{\partial #2}}
\newcommand{\diff}[2]{\frac{d #1}{d #2}}

\newcommand{\myeqref}[1]{Eq.~(\ref{#1})}
\newcommand{\myfigref}[1]{Fig.~\ref{#1}}

\usepackage{url}
\usepackage{pgfplots}
\usepackage{algorithm,algpseudocode}
\usepackage{algorithmicx}
\usepackage{amsmath}
\usepackage{amsfonts}
\usepackage{gensymb}
\usepackage{siunitx}\usepackage[colorlinks,bookmarksnumbered,bookmarksopen,citecolor=blue,urlcolor=blue,linkcolor=blue,]{hyperref}
\usepackage{xcolor}

\usepackage{pgfplotstable}
\usepgfplotslibrary{statistics}
\pgfplotsset{
/pgfplots/custom legend/.style={
legend image code/.code={
\draw [only marks,mark=square]
plot coordinates {(0.3cm,0cm)};
}, },}

\makeatletter
\pgfplotsset{
    boxplot prepared from table/.code={
        \def\tikz@plot@handler{\pgfplotsplothandlerboxplotprepared}%
        \pgfplotsset{
            /pgfplots/boxplot prepared from table/.cd,
            #1,
        }
    },
    /pgfplots/boxplot prepared from table/.cd,
        table/.code={\pgfplotstablecopy{#1}\to\boxplot@datatable},
        row/.initial=0,
        make style readable from table/.style={
            #1/.code={
                \pgfplotstablegetelem{\pgfkeysvalueof{/pgfplots/boxplot prepared from table/row}}{##1}\of\boxplot@datatable
                \pgfplotsset{boxplot/#1/.expand once={\pgfplotsretval}}
            }
        },
        make style readable from table=lower whisker,
        make style readable from table=upper whisker,
        make style readable from table=lower quartile,
        make style readable from table=upper quartile,
        make style readable from table=median,
        make style readable from table=lower notch,
        make style readable from table=upper notch
}
\makeatother

\title{Thermodynamics-informed Graph Neural Networks} 

\author[1]{Quercus Hern\'andez}
\author[2]{Alberto Bad\'ias}
\author[3]{Francisco Chinesta}
\author[1]{El\'ias Cueto}
\affil[1]{{\small Aragon Institute of Engineering Research (I3A). University of Zaragoza. Zaragoza, Spain.}}
\affil[2]{{\small Higher Technical School of Industrial Engineering, Polytechnic University of Madrid. Madrid, Spain.}}
\affil[3]{{\small ESI Group chair. PIMM Lab. ENSAM Institute of Technology. Paris, France. CNRS@CREATE LTD. Singapore.}}

\title{Thermodynamics-informed Graph Neural Networks} 

\begin{document} 

\maketitle

\begin{abstract}
In this paper we present a deep learning method to predict the temporal evolution of dissipative dynamic systems. We propose using both geometric and thermodynamic inductive biases to improve accuracy and generalization of the resulting integration scheme. The first is achieved with Graph Neural Networks, which induces a non-Euclidean geometrical prior with permutation invariant node and edge update functions. The second bias is forced by learning the GENERIC structure of the problem, an extension of the Hamiltonian formalism, to model more general non-conservative dynamics. Several examples are provided in both Eulerian and Lagrangian description in the context of fluid and solid mechanics respectively, achieving relative mean errors of less than 3\% in all the tested examples. Two ablation studies are provided based on recent works in both physics-informed and geometric deep learning.
\end{abstract}

\section{Introduction}

Computer simulation has become a standard tool in a wide variety of fields, from natural to social sciences, in order to simulate physical phenomena and predict its future behaviour. Simulation models enable engineers to find optimal designs which can later be validated in a much more expensive experimental set-ups. Thus, it is important to develop models based on mathematical and physical foundations which ensure the reliability of the results. With the irruption of the information era, recent research lines have focused on the use of data-based deep learning algorithms which overcome some of the limitations of traditional methods, such as handling highly nonlinear dynamics \cite{raissi2019physics}, problems whose mathematical formulation is unknown \cite{cranmer2020discovering} or real-time simulation performance \cite{moya2021physics}.

However, deep learning is usually very demanding in both data and computational power. A present challenge is not only to develop numerical algorithms that speed-up the already complicated neural network models but also to build smarter architectures which take advantage of the problem structure, potentially reducing the data consumption and improving its generalization. Two historical examples are convolutional \cite{lecun1999object} and recurrent \cite{rumelhart1985learning} neural networks, which exploit the structured data in grid elements and timesteps to induce translation and time invariance respectively, both remaining as major breakthroughs in their respective fields. 

Motivated by the study of such architectures, which exploit the invariant quantities of the problem, a new machine learning paradigm recently arose known as geometric deep learning \cite{bronstein2017geometric}. The key insight is to impose specific constraints related to the symmetries of the problem, acting as a strong inductive bias to the learning process. This framework is not only restricted to regular structured data (images or time series) but also to a general case of relations (edges) over arbitrary entities (nodes), mathematically represented with a graph.

In a similar way, physics problems have an intrinsic mathematical structure unveiled by centuries of theoretical and experimental knowledge. From Newton's laws of motion to the Standard Model, the laws of nature can often be described as a set of partial differential equations (PDEs) which can also lead neural networks to find the correct solution of a dynamical problem. These are the foundations of the so-called physics-informed deep learning.

The present work aims to combine both inductive biases in order to learn a physical simulator able to predict the dynamics of complex systems in the context of fluid and solid mechanics.

\section{Background}

\subsection{Physics-informed deep learning}

Recent works about predicting physics with neural networks \cite{raissi2018hidden,raissi2019physics} have demonstrated the convenience of using physical constraints to exploit the dynamical properties of a simulation problem. In fact, the Lagrangian and Hamiltonian reformulations of Newton's classical equations of motion describe a rich mathematical theory based on the Poisson bracket and a symplectic manifold structure. This topological property can be applied to neural networks obtaining robust time integrators, as shown in plenty of recent literature \cite{greydanus2019hamiltonian,sanchez2019hamiltonian,mattheakis2020hamiltonian,desai2021port,chen2019symplectic}. 

This structure can be extended to non-conservative systems using a dissipative bracket with the so-called GENERIC formalism \cite{ottinger1997dynamics, grmela1997dynamics}. Several works have already addressed the imposition of this structure bias \cite{hernandez2021structure,hernandez2021deep,lee2021machine,moya2021physics,zhang2021gfinns}, but are hard to learn as the standard multilayer perceptron architectures may not be adequate for larger systems or phenomena whose GENERIC formulation is more complex.

\subsection{Geometric deep learning}

The consideration of a general graph theory framework is motivated by the fact that certain data are not structured in the usual Euclidean space, but in a more complex manifold. For instance, solving PDEs on arbitrary domains is a common problem in physics and engineering, whose solution is rarely found in an Euclidean manifold.

Several graph-based works have achieved great improvements in physics problems such as predicting atomic bond forces \cite{hu2021forcenet}, particle tracking in high energy physics \cite{shlomi2020graph,ju2020graph}, n-body problem with more general interactions \cite{kipf2018neural,battaglia2018relational,cranmer2020discovering}, or learning simulators to predict complex fluid interactions \cite{sanchez2020learning} and meshed domains \cite{pfaff2020learning,zheng2021deep}. These last approaches include the dynamics of the system by predicting the velocity or acceleration, depending on the order of the governing PDE, and integrating them in time. However, they do not include any physical information about the rest of the predicted variables, which remain as black-box direct predictions. 

In this work, we propose a learning method of the GENERIC operators and potentials using graph neural networks. This graph-based architecture exploits both the geometrical constraints of discretized systems and the thermodynamically consistent integration scheme of the GENERIC formalism to predict the time evolution of dynamical systems. 

The present paper is organized as follows. In section III, we formalize the problem statement and introduce the metriplectic and geometrical inductive biases induced by the proposed architecture. In section IV, we provide three examples of fluid and solid mechanics comparing the results with other methods. Concluding remarks are given in section V.

\section{Methodology}\label{sec:method}

\subsection{Problem Statement}

Let $\bs z \in \mathcal M \subseteq \mathbb R^n$ be the independent state variables which fully describe a dynamical system up to a certain level of description, with $\mathcal M$ the set of all the admissible states (space state) which is assumed to have the structure of a differentiable manifold in $\mathbb R^n$. The given physical phenomenon can be expressed as a system of differential equations encoding the time evolution of its state variables $\bs z$,
\begin{equation}\label{eq:PDE}
\dot{\bs z}= \diff{\bs{z}}{t} = F(\bs{z},t),\; t\in\mathcal{I}=(0,T],\; \bs z(0)=\bs z_0,
\end{equation}
where $t$ refers to the time coordinate in the time interval $\mathcal{I}$ and {$F(\bs z, t)$} refers to an arbitrary nonlinear differential function.

The goal of this paper is to find the convenient mapping $F$ for a dynamical system governed by \myeqref{eq:PDE} from data, in order to efficiently predict its time evolution after a prescribed time horizon $T$ using deep learning \cite{weinan2017proposal}. The solution is forced to fulfill the basic thermodynamic requirements of energy conservation and entropy inequality restrictions via the GENERIC formalism.

\subsection{Metriplectic structure: The GENERIC formalism}

In this work, we guarantee the physical meaning of the solution by enforcing the GENERIC structure of the system \cite{ottinger1997dynamics, grmela1997dynamics}, a generalization of the Hamiltonian paradigm to dissipative systems. The GENERIC formulation of time evolution for nonequilibrium systems, described by the set of $\bs{z}$ state variables, is given by
\begin{equation}\label{eq:brackets}
\frac{d\bs{z}}{dt}=\lbrace\bs{z},E\rbrace+[\bs{z},S].
\end{equation}
The conservative contribution is assumed to be of Hamiltonian form, requiring an energy potential $E=E(\bs{z})$ and a Poisson bracket $\lbrace\bs{z},E\rbrace$ acting on the state vector. This accounts for the reversible dynamics in classical mechanics, expressed by the time structure invariance of the Poisson bracket. Similarly, the remaining irreversible contribution to the energetic balance of the system is generated by the nonequilibrium entropy potential $S=S(\bs{z})$ with an irreversible or friction bracket $[\bs{z},S]$. This term is considered an extension of Landau's idea of the time evolution of a state variable towards its equilibrium value via a dissipation potential \cite{landau1965collected}.

For practical use, it is convenient to reformulate the bracket notation using two linear operators:
\begin{equation}
\bs L : T^*\mathcal M \rightarrow T\mathcal M,\quad \bs M : T^*\mathcal M \rightarrow T\mathcal M,
\end{equation}
where $T^*\mathcal M$ and $T\mathcal M$ represent, respectively, the cotangent and tangent bundles of the state space $\mathcal M$. These operators inherit the mathematical properties of the original bracket formulation. The operator $\bs{L}=\bs{L}(\bs{z})$ represents the Poisson bracket and is required to be skew-symmetric (a cosymplectic matrix). Similarly, the friction matrix $\bs{M}=\bs{M}(\bs{z})$ accounts for the irreversible part of the system and is symmetric and positive semi-definite to ensure the positiveness of the dissipation rate. 

Replacing the original bracket formulation in \myeqref{eq:brackets} with their respective operators, the time-evolution equation for the state variables $\bs{z}$ is derived,
\begin{equation}\label{eq:generic}
\diff{\bs{z}}{t} = \bs{L} \dpar{E}{\bs{z}} + \bs{M} \dpar{S}{\bs{z}}.
\end{equation}

However, the restrictions of $\bs{L}$ and $\bs{M}$ are not sufficient to guarantee a thermodynamically consistent description of the dynamics of the system, and two degeneracy conditions are introduced 
\begin{equation}
\lbrace S,\bs{z}\rbrace=[ E,\bs{z}]=\bs{0}.
\end{equation}
The first one states that the entropy is a degenerate functional of the Poisson bracket, showing the reversible nature of the Hamiltonian contribution to the dynamics. The second expression states the conservation of the total energy of the system with a degenerate condition of the energy with respect to the friction bracket. These restrictions can be reformulated in a matrix form in terms of the $\bs{L}$ and $\bs{M}$ operators, resulting in the following degeneracy restrictions:
\begin{equation}\label{eq:degen}
\bs{L} \dpar{S}{\bs{z}}=\bs{M} \dpar{E}{\bs{z}} = \bs 0.
\end{equation}

The degeneracy conditions, in addition to the non-negativeness of the irreversible bracket, guarantees the first (energy conservation) and second (entropy inequality) laws of thermodynamics,

\begin{equation}\label{eq:Econs}
\frac{dE}{dt}=\lbrace E,E\rbrace =0,\quad\frac{dS}{dt}=[ S,S]\geq 0.
\end{equation}

The specific topological requirements based on a pair of skew-symmetric ($\bs{L}$) and symmetric ($\bs{M}$) operators over a smooth manifold is called a metriplectic structure \cite{morrison1986paradigm, guha2007metriplectic}. We make use of this hard constraint to construct a thermodynamically-sound integrator, acting as the first inductive bias of our approach. This integration is performed using a forward Euler scheme with time increment $\Delta t$ and the GENERIC formalism in \myeqref{eq:generic}, resulting in the following expression
\begin{equation}
\label{eq:generic_discrete}
\bs{z}_{t+1}=\bs{z}_t+\Delta t\left(\bs{L}\dpar{E}{\bs{z}_t} + \bs{M} \dpar{S}{\bs{z}_t}\right).
\end{equation}

In order to learn the GENERIC operators $\bs{L}$, $\bs{M}$ and potentials $E$, $S$ for each particle of the domain, we propose the use of graph-based deep learning, which exploits the geometrical structure of that specific domain.

\subsection{Geometric structure: Graph Neural Networks}

Let $\mathcal{G}=(\mathcal{V},\mathcal{E},\bs{u})$ be a directed graph, where $\mathcal{V}=\{1,...,n\}$ is a set of $|\mathcal{V}|=n$ vertices, $\mathcal{E}\subseteq\mathcal{V}\times\mathcal{V}$ is a set of $|\mathcal{E}|=e$ edges and $\bs{u}\in\mathbb{R}^{F_g}$ is the global feature vector. Each vertex and edge in the graph is associated with a node and a pairwise interaction between nodes respectively in a discretized physical system. The global feature vector defines the properties shared by all the nodes in the graph, such as gravity or elastic properties.  For each vertex $i\in\mathcal{V}$ we associate a feature vector $\bs{v}_i\in\mathbb{R}^{F_v}$, which represents the physical properties of each individual node. Similarly, for each edge $(i,j)\in\mathcal{E}$ we associate an edge feature vector $\bs{e}_{ij}\in\mathbb{R}^{F_e}$. 

\begin{figure}[h]
\centering
\includegraphics[width=0.25\textwidth]{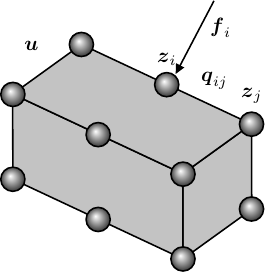}
\hspace{0.1\textwidth}
\includegraphics[width=0.25\textwidth]{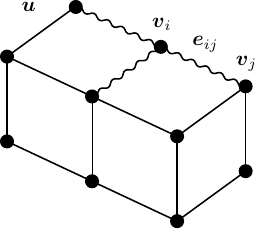}
\caption{Physical system domain discretized in a mesh with node state variables $\bs{z}_i$, relative nodal distances $\bs{q}_{ij}$, external interactions $\bs{f}_i$ and global properties $\bs{u}$ (top). Graph representation of the same system, with node and edge attributes: $\bs{v}_i$ and $\bs{e}_{ij}$ (bottom).\label{fig:graphnet}}
\end{figure}

In practice, the positional state variables of the system ($\bs{q}_i$) are assigned to the edge feature vector $\bs{e}_{ij}$ so the edge features represent relative distances ($\bs{q}_{ij}=\bs{q}_i-\bs{q}_j$) between nodes, giving a distance-based attentional flavour to the graph network \cite{monti2018dual,velivckovic2017graph,zhang2018gaan} and translational invariance \cite{wang2019dynamic,shi2020point}. The rest of the state variables are assigned to the node feature vector $\bs{v}_{i}$. The external interactions, such as forces applied to the system, are included in an external load vector $\bs{f}_i$. A simplified scheme of the graph codification of a physical system is depicted in \myfigref{fig:graphnet}.

\begin{figure*}[h]
\includegraphics[width=\textwidth]{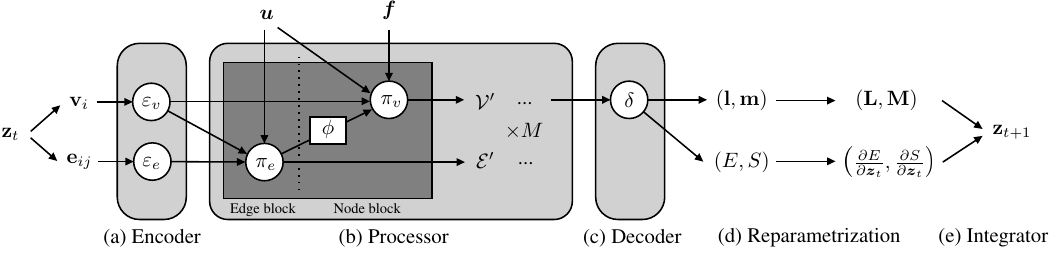}
\caption{Algorithm block scheme used to predict a single-step state variable change in time. (a) The encoder transforms the node and edge features to a learnt embedding. (b) The processor shares the nodal information through the graph via $M$ message passing modules. (c) The decoder extracts the GENERIC flattened operators and potentials from the processed node embeddings. (d) The reparametrization step builds the symmetries of the $\bs{L}$ and $\bs{M}$ operators and computes the potential gradients with respect to the network input. (e) The integrator predicts the next time step state variables based on the GENERIC formulation. The whole process is repeated iteratively to get the dynamical rollout of the physical system.\label{fig:algorithm}}
\end{figure*}

These features are fed into an encode-process-decode scheme \cite{battaglia2018relational}, which consists on several multilayer perceptrons (MLPs) shared between all the nodes and edges of the graph. The algorithm consists of five steps (\myfigref{fig:algorithm}): 

\subsubsection{Encoding}

We use two MLPs ($\varepsilon_v$, $\varepsilon_e$) to transform the vertex and edge initial feature vectors into higher-dimensional embeddings $\bs{x}_i\in\mathbb{R}^{F_h}$ and $\bs{x}_{ij}\in\mathbb{R}^{F_h}$ respectively,
\begin{align}
\begin{split}
\varepsilon_e:\mathbb{R}^{F_e} & \longrightarrow \mathbb{R}^{F_h} \\
\bs{e}_{ij} & \longmapsto \bs{x}_{ij} 
\end{split}
\begin{split}
\varepsilon_v:\mathbb{R}^{F_v} & \longrightarrow \mathbb{R}^{F_h} \\
\bs{v}_i & \longmapsto \bs{x}_i .
\end{split}
\end{align}

\subsubsection{Processing}

The processor is the core task of the algorithm, as it shares the nodal information between vertices via message passing and modifies the hidden vectors in order to extract the desired output of the system. First, a MLP ($\pi_e$) computes the updated edge features $\bs{x}'_{ij}$ for each graph edge, based on the current edge features, global features, and sending and recieving node, 
\begin{align}
\begin{split}
\pi_e:\mathbb{R}^{3F_h+F_g} & \longrightarrow \mathbb{R}^{F_h}\\
(\bs{x}_{ij},\bs{x}_i,\bs{x}_j,\bs{u}) & \longmapsto \bs{x}'_{ij}.
\end{split}
\end{align}
Then, for each node the messages are pooled with a permutation invariant function $\phi$ based on the neighborhood $\mathcal{N}_i=\{j\in\mathcal{V}|(i,j)\in\mathcal{E}\}$ of the node $i$. Last, the node embeddings are updated with a second MLP ($\pi_v$) using the current node features, the pooled messages, the external load vector and the global features,
\begin{align}
\begin{split}
\pi_v:\mathbb{R}^{2F_h+F_f+F_g} & \longrightarrow \mathbb{R}^{F_h}\\
(\bs{x}_i,\phi(\bs{x}'_{ij}),\bs{f}_i,\bs{u}) & \longmapsto \bs{x}'_i
\end{split}
\end{align}
where $(\cdot,\cdot)$ denotes vector concatenation and $\bs{x}'_i$ and $\bs{x}'_{ij}$ are the updated nodal and edge latent vectors.

The processing step is equivalent to the message passing \cite{gilmer2017neural} of 1-step adjacent nodes. In order to get the influence of further graph nodes, the process can be recurrently repeated with both shared or unshared parameters in $M$ processing blocks and optionally using residual connections \cite{he2016deep}. In this approach, we use both unshared parameters and residual connections to each message passing block and sum as aggregation function $\phi$. Note that the computed messages $\phi(\bs{x}'_{ij})$ represent a hidden embedding of the intermolecular interactions of the system (internal messages) whereas the vector $\bs{f}_i$ accounts for the external interactions (external messages).

\subsubsection{Decoding}

The last block extracts the relevant physical output information $\bs{y}_i\in\mathbb{R}^{F_y}$ of the system from the node latent feature vector, implemented with a MLP ($\delta_v$). In this work, we predict for each particle the GENERIC energy $E$ and entropy $S$ potentials and the flattened operators $\bs{l}$ and $\bs{m}$:
\begin{align}
\begin{split}
\delta_v:\mathbb{R}^{F_h} & \longrightarrow \mathbb{R}^{F_y} \\
\bs{x}'_i & \longmapsto \bs{y}_i=(\bs{l},\bs{m},E,S).
\end{split}
\end{align}

\subsubsection{Reparametrization}

A last processing step is needed to get the GENERIC parameters before integrating the state variables. Both operators in matrix form $\bs{L}$ and $\bs{M}$ are constructed using the flattened output of the Graph Neural Network $\bs{l}$ and $\bs{m}$ respectively, reshaped in lower-triangular matrices. The skew-symmetric and positive semi-definite conditions are imposed by construction using the following parametrization:
\begin{equation}
\label{eq:gnn_output}
\bs{L}=\bs{l}-\bs{l}^\top, \qquad \bs{M}=\bs{m}\bs{m}^\top. 
\end{equation}

Both $E$ and $S$ are directly predicted for every node. Then, these potentials can be differentiated with respect to the network input in order to get the gradients $\dpar{E}{\bs{z}}$ and $\dpar{S}{\bs{z}}$ needed for the GENERIC integrator. These gradients are easily obtained using automatic differentiation \cite{paszke2017automatic}, and ensures the integrability of the energy and entropy gradients \cite{teichert2019machine}.

Considering the dimensions of the lower triangular matrices and the scalar value of both potentials, the output dimension of the decoder network is 
\begin{equation}
F_y=\frac{n(n+1)}{2}+\frac{n(n-1)}{2}+1+1
\end{equation}
where $n$ represent the dimension of the state variables $\bs{z}$.

\subsubsection{Integration}

The single-step integration of the state variables of the system $\bs{z}_t\rightarrow\bs{z}_{t+1}$ is then performed using \myeqref{eq:generic_discrete}.

\subsection{Learning procedure}

The complete dataset $\mathcal{D}$ is composed by $N_\text{sim}$ multiparametric simulation cases of a dynamical system evolving in time. Each case $\mathcal{D}_i$ contains the labelled pair of a single-step state vector $\bs{z}_t$ and its evolution in time $\bs{z}_{t+1}$ for each node of the system
\begin{equation}
\mathcal{D}=\{\mathcal{D}_i\}_{i=1}^{N_{\text{sim}}}, \quad
\mathcal{D}_i =\{(\bs{z}_t,\bs{z}_{t+1})\}_{t=0}^{T},
\end{equation}
where the dataset $\mathcal{D}$ is disjointly partitioned in $80\%$ training, $10\%$ test and $10\%$ validation sets: $\mathcal{D}_{\text{train}}$, $\mathcal{D}_{\text{val}}$ and $\mathcal{D}_{\text{test}}$ respectively. 

The training is performed in a single-snapshot supervision, which has two main advantages: (i) enables parallelization between snapshots, which decreases training time, and (ii) avoids intensive memory usage due to a several-snapshot recursive training. The loss function is divided in two terms:

\begin{itemize}
\item Data loss: This term accounts for the correct prediction of the state vector time evolution using the GENERIC integrator. It is defined as the MSE along the graph nodes and state variables between the predicted and the ground-truth time derivative of the state vector in a given snapshot, 
\begin{equation}
\mathcal{L}^{\text{data}}_n=\left\Vert\frac{d\bs{z}^{\text{GT}}}{dt}-\frac{d\bs{z}^{\text{net}}}{dt}\right\Vert^2_2,
\end{equation}
where $\Vert\cdot\Vert_2$ denotes the L2-norm. The choice of the time derivative instead of the state vector itself is to regularize the global loss function to a uniform order of magnitude with respect to the degeneracy terms, as shown in \myeqref{eq:generic_discrete}.

\item Degeneracy loss: This condition is added to the optimization in order to force the degeneracy conditions of the Poisson and dissipative operators, which ensure thermodynamical consistency of the integrator. It is defined as the MSE along the graph nodes and state variables of two residual terms corresponding to the energy and entropy degeneracy conditions,
\begin{equation}
\mathcal{L}^{\text{deg}}_n=\left\Vert\bs{L}\dpar{S}{\bs{z}_n}\right\Vert^2_2+\left\Vert\bs{M}\dpar{E}{\bs{z}_n}\right\Vert^2_2.
\end{equation}
Alternative approaches are found in the literature to impose this degeneracy restrictions, such as a specific tensor parametrization of the brackets \cite{lee2021machine} or forcing ortogonality using additional skew-symmetric matrices \cite{zhang2021gfinns}. However, we decide to include it as a soft constraint in order to allow more flexibility in the learning process and improve convergence while maintaining the degeneracy conditions up to an admissible error.
\end{itemize}

The global loss term is a weighted mean of the two terms over the shuffled $N_{\text{batch}}$ batched snapshots,
\begin{equation}
\mathcal{L}=\frac{1}{N_{\text{batch}}}\sum_{n=0}^{N_{\text{batch}}}(\lambda \mathcal{L}^{\text{data}}_n+\mathcal{L}^{\text{deg}}_n).
\end{equation}

As the energy and entropy are supervised only by their gradients, we remark that (i) they are learnt up to an integration constant value and (ii) the activation functions must have a sufficient degree of continuity. To meet this second requirement, one must select activations with non-zero second derivative in order to have a correct backpropagation of the weights and biases. Thus, linear or rectified units (ReLU, Leaky ReLU, RReLU) are not appropriate for this task. It is well known \cite{hornik1991approximation} that logistic functions such as sigmoid and hyperbolic tangent are universal approximators of any derivative arbitrarily well, but are not optimal for very deep neural networks architectures, as they suffer from several problems such as vanishing gradients. Then, the correct activation functions suitable for learning gradients are the ones which combine both non-zero second derivatives and ReLU-type non-linearities, such as Softplus, Swish \cite{ramachandran2017searching} or Mish \cite{misra2019mish}. In the present work we use the Swish activation function.

The inputs and outputs of the networks are standardized using the training dataset statistics. Gaussian noise is also added to the inputs during training in order to model the accumulation of error during the time integration \cite{pfaff2020learning}, which is not contemplated in a single-snapshot training, with the variance of the noise $\sigma_\text{noise}^2$ as a tunable hyperparameter and zero mean value. All the cases are optimized using Adam \cite{kingma2014adam} and a multistep learning rate scheduler.

The code is fully implemented in Pytorch. Our datasets and trained networks are publicly available online at \url{https://github.com/quercushernandez}.

\subsection{Evaluation metrics}

Two ablation studies are performed to evaluate the method presented in this work. The first case is performed using only Graph Neural Networks (from now, GNN) with similar architecture and learning procedure used in prior works \cite{sanchez2020learning,pfaff2020learning} and no metriplectic integrator. In the second case, we impose the metriplectic structure \cite{hernandez2021structure,hernandez2021deep} (from now, SPNN), using standard MLPs with no graph computations. Both alternative methods are tuned for equal parameter count in order to get a fair comparison of the results.

All the results are computed with the integration scheme in \myeqref{eq:generic_discrete} iteratively from the initial conditions to the prescribed time horizon $T$, denoted as rollout. The rollout prediction error is quantified by the relative L2 error, computed with \myeqref{eq:L2error} for each snapshot and simulation case,
\begin{equation}
\label{eq:L2error}
\varepsilon=\frac{\Vert\bs{z}^\text{GT}-\bs{z}^\text{net}\Vert_2}{\Vert\bs{z}^\text{GT}\Vert_2}.
\end{equation}
The results are represented in Fig. \ref{fig:couette_box}, \ref{fig:beam_box} and \ref{fig:cyl_box} showing the rollout statistics for all the snapshots divided in train and test simulations, state variables and method used (Ours, GNN or SPNN).

\section{Numerical experiments}

\subsection{Couette flow of an Oldroyd-B fluid}\label{sec:couette}

\subsubsection{Description}

\begin{figure}[h]
\centering
\includegraphics[width=0.45\textwidth]{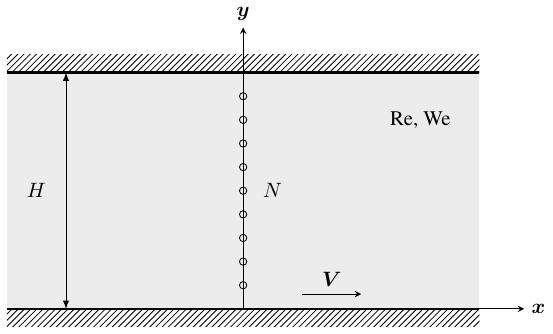}  
\caption{Couette flow in an Oldroyd-B fluid. The simulations span different Reynolds and Weissenberg numbers to obtain different flow profiles with a fixed lid velocity.\label{fig:couette}}
\end{figure}

The first example is a shear (Couette) flow of an Oldroyd-B fluid model (\myfigref{fig:couette}). This is a constitutive model for viscoelastic fluids, considering linear elastic dumbbells as a proxy representation of polymeric chains immersed in a solvent.

The state variables chosen are the position of the fluid on each node of the mesh $\bs{q}$, its velocity $v$ in the $x$ direction, internal energy $e$ and the conformation tensor shear component $\tau$,
\begin{equation}\label{eq:couette_z}
\mathcal{S}=\{\bs{z}=(\bs{q}, v,e,\tau)\in\mathbb{R}^2\times\mathbb{R}\times\mathbb{R}\times\mathbb{R}\}.
\end{equation}
The edge feature vector contains the relative position of the nodes whereas the rest of the state variables are part of the node feature vector. An additional one-hot vector $\bs{n}$ is added to the node features in order to represent the boundary and fluid nodes. The global feature vector $\bs{u}$ represent the Weissenberg and Reynolds numbers of each simulation, resulting in the following feature vectors:
\begin{equation}\label{eq:couette_feat}
\bs{e}_{ij}=(\bs{q}_i-\bs{q}_j,\Vert\bs{q}_{i}-\bs{q}_{j}\Vert_2),\quad\bs{v}_i=(v,e,\tau,\bs{n}),\quad\bs{u}=(\text{Re},\text{We}).
\end{equation}

\subsubsection{Database and Hyperparameters}

The training database for the Couette flow is generated with the CONNFFESSIT technique \cite{laso1993calculation}, based on the Fokker-Plank equation \cite{le2009multiscale}, using a Monte Carlo algorithm. The fluid is discretized in the vertical direction with $N_e=100$ elements and $N=101$ nodes in a total height of $H=1$. A total of 10,000 dumbells are considered at each nodal location in the model. The lid velocity is set to $V=1$, with variable Weissenberg $\text{We}\in[1,2]$ and Reynolds number $\text{Re}\in[0.1,1]$, summing a total of $N_\text{sim}=100$ cases. The simulation is discretized in $N_T=150$ time increments of $\Delta t = 6.7\e{-3}$.

Following \myeqref{eq:couette_feat}, the dimensions of the graph feature vectors are $F_e=3$, $F_v=5$ and $F_g=2$. The hidden dimension of the node and edge latent vectors is $F_h=10$. The learning rate is set to $l_r=10^{-3}$ with decreasing order of magnitude on epochs $2000$ and $4000$, and a total number of $N_\text{epoch}=6000$. The training noise variance is set to $\sigma_\text{noise}^2=10^{-2}$.

\subsubsection{Results}

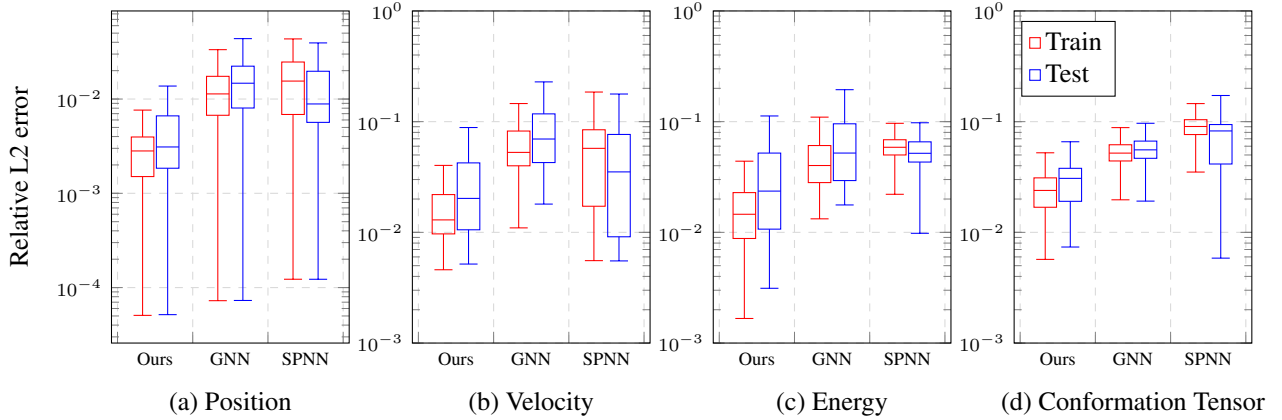
\begin{figure*}[h]
\centering
\pgfplotstableread{results/couette_box/mse_ours_train.txt}\mseourstrain
\pgfplotstableread{results/couette_box/mse_ours_test.txt}\mseourstest
\pgfplotstableread{results/couette_box/mse_gnn_train.txt}\msegnntrain
\pgfplotstableread{results/couette_box/mse_gnn_test.txt}\msegnntest
\pgfplotstableread{results/couette_box/mse_spnn_train.txt}\msespnntrain
\pgfplotstableread{results/couette_box/mse_spnn_test.txt}\msespnntest

\begin{tikzpicture}
\pgfplotsset{width=\textwidth, height=6cm}

  \begin{semilogyaxis}
  [xshift=0cm,xlabel=(a) Position,
  area legend, boxplot/draw direction=y,
  grid=major, 
  grid style={dashed,gray!30}, 
  cycle list={{red},{blue}},
  boxplot={draw position={1/3 + floor(\plotnumofactualtype/2) + 1/3*mod(\plotnumofactualtype,2)},box extend=0.3,},
  x=1cm,xtick={0,1,2,...,10},x tick label as interval,
  xticklabels={{Ours},{GNN},{SPNN}},
  ticklabel style={font=\scriptsize},
  ylabel={Relative L2 error},
  ] 
	\addplot+[boxplot prepared from table={table=\mseourstrain,row=0,
    lower whisker=lw,
    upper whisker=uw,
    lower quartile=lq,
    upper quartile=uq,
    median=med}, boxplot prepared] 
    coordinates {};
	\addplot+[boxplot prepared from table={table=\mseourstest,row=0,
    lower whisker=lw,
    upper whisker=uw,
    lower quartile=lq,
    upper quartile=uq,
    median=med}, boxplot prepared] 
    coordinates {}; 
    
	\addplot+[boxplot prepared from table={table=\msegnntrain,row=0,
    lower whisker=lw,
    upper whisker=uw,
    lower quartile=lq,
    upper quartile=uq,
    median=med}, boxplot prepared] 
    coordinates {}; 
	\addplot+[boxplot prepared from table={table=\msegnntest,row=0,
    lower whisker=lw,
    upper whisker=uw,
    lower quartile=lq,
    upper quartile=uq,
    median=med}, boxplot prepared] 
    coordinates {}; 
        
	\addplot+[boxplot prepared from table={table=\msespnntrain,row=0,
    lower whisker=lw,
    upper whisker=uw,
    lower quartile=lq,
    upper quartile=uq,
    median=med}, boxplot prepared] 
    coordinates {}; 
	\addplot+[boxplot prepared from table={table=\msespnntest,row=0,
    lower whisker=lw,
    upper whisker=uw,
    lower quartile=lq,
    upper quartile=uq,
    median=med}, boxplot prepared] 
    coordinates {};
    
  \end{semilogyaxis}
  
  \begin{semilogyaxis}
  [xshift=4cm,xlabel=(b) Velocity,
  area legend, boxplot/draw direction=y,
  grid=major, 
  grid style={dashed,gray!30}, 
  cycle list={{red},{blue}},
  boxplot={draw position={1/3 + floor(\plotnumofactualtype/2) + 1/3*mod(\plotnumofactualtype,2)},box extend=0.3,},
  x=1cm,xtick={0,1,2,...,10},x tick label as interval,
  xticklabels={{Ours},{GNN},{SPNN}},
  ticklabel style={font=\scriptsize},
  ymin=0.001,ymax=1,
  ] 
	\addplot+[boxplot prepared from table={table=\mseourstrain,row=1,
    lower whisker=lw,
    upper whisker=uw,
    lower quartile=lq,
    upper quartile=uq,
    median=med}, boxplot prepared] 
    coordinates {};
	\addplot+[boxplot prepared from table={table=\mseourstest,row=1,
    lower whisker=lw,
    upper whisker=uw,
    lower quartile=lq,
    upper quartile=uq,
    median=med}, boxplot prepared] 
    coordinates {}; 
    
	\addplot+[boxplot prepared from table={table=\msegnntrain,row=1,
    lower whisker=lw,
    upper whisker=uw,
    lower quartile=lq,
    upper quartile=uq,
    median=med}, boxplot prepared] 
    coordinates {}; 
	\addplot+[boxplot prepared from table={table=\msegnntest,row=1,
    lower whisker=lw,
    upper whisker=uw,
    lower quartile=lq,
    upper quartile=uq,
    median=med}, boxplot prepared] 
    coordinates {}; 
        
	\addplot+[boxplot prepared from table={table=\msespnntrain,row=1,
    lower whisker=lw,
    upper whisker=uw,
    lower quartile=lq,
    upper quartile=uq,
    median=med}, boxplot prepared] 
    coordinates {}; 
	\addplot+[boxplot prepared from table={table=\msespnntest,row=1,
    lower whisker=lw,
    upper whisker=uw,
    lower quartile=lq,
    upper quartile=uq,
    median=med}, boxplot prepared] 
    coordinates {};
    
  \end{semilogyaxis}

  \begin{semilogyaxis}
  [xshift=8cm,xlabel=(c) Energy,
  area legend, boxplot/draw direction=y,
  grid=major, 
  grid style={dashed,gray!30}, 
  cycle list={{red},{blue}},
  boxplot={draw position={1/3 + floor(\plotnumofactualtype/2) + 1/3*mod(\plotnumofactualtype,2)},box extend=0.3,},
  x=1cm,xtick={0,1,2,...,10},x tick label as interval,
  xticklabels={{Ours},{GNN},{SPNN}},
  ticklabel style={font=\scriptsize},
  ymin=0.001,ymax=1,
  ] 
	\addplot+[boxplot prepared from table={table=\mseourstrain,row=2,
    lower whisker=lw,
    upper whisker=uw,
    lower quartile=lq,
    upper quartile=uq,
    median=med}, boxplot prepared] 
    coordinates {};
	\addplot+[boxplot prepared from table={table=\mseourstest,row=2,
    lower whisker=lw,
    upper whisker=uw,
    lower quartile=lq,
    upper quartile=uq,
    median=med}, boxplot prepared] 
    coordinates {}; 
    
	\addplot+[boxplot prepared from table={table=\msegnntrain,row=2,
    lower whisker=lw,
    upper whisker=uw,
    lower quartile=lq,
    upper quartile=uq,
    median=med}, boxplot prepared] 
    coordinates {}; 
	\addplot+[boxplot prepared from table={table=\msegnntest,row=2,
    lower whisker=lw,
    upper whisker=uw,
    lower quartile=lq,
    upper quartile=uq,
    median=med}, boxplot prepared] 
    coordinates {}; 
        
	\addplot+[boxplot prepared from table={table=\msespnntrain,row=2,
    lower whisker=lw,
    upper whisker=uw,
    lower quartile=lq,
    upper quartile=uq,
    median=med}, boxplot prepared] 
    coordinates {}; 
	\addplot+[boxplot prepared from table={table=\msespnntest,row=2,
    lower whisker=lw,
    upper whisker=uw,
    lower quartile=lq,
    upper quartile=uq,
    median=med}, boxplot prepared] 
    coordinates {};
    
  \end{semilogyaxis}
 
  \begin{semilogyaxis}
  [xshift=12cm,xlabel=(d) Conformation Tensor,
  area legend, boxplot/draw direction=y,
  grid=major, 
  grid style={dashed,gray!30}, 
  cycle list={{red},{blue}},
  boxplot={draw position={1/3 + floor(\plotnumofactualtype/2) + 1/3*mod(\plotnumofactualtype,2)},box extend=0.3,},
  x=1cm,xtick={0,1,2,...,10},x tick label as interval,
  xticklabels={{Ours},{GNN},{SPNN}},
  ticklabel style={font=\scriptsize},
  ymin=0.001, ymax=1,
  custom legend,legend pos=north west,legend cell align=left,
  legend entries = {Train, Test},
  ] 
	\addplot+[boxplot prepared from table={table=\mseourstrain,row=3,
    lower whisker=lw,
    upper whisker=uw,
    lower quartile=lq,
    upper quartile=uq,
    median=med}, boxplot prepared] 
    coordinates {};
	\addplot+[boxplot prepared from table={table=\mseourstest,row=3,
    lower whisker=lw,
    upper whisker=uw,
    lower quartile=lq,
    upper quartile=uq,
    median=med}, boxplot prepared] 
    coordinates {}; 
    
	\addplot+[boxplot prepared from table={table=\msegnntrain,row=3,
    lower whisker=lw,
    upper whisker=uw,
    lower quartile=lq,
    upper quartile=uq,
    median=med}, boxplot prepared] 
    coordinates {}; 
	\addplot+[boxplot prepared from table={table=\msegnntest,row=3,
    lower whisker=lw,
    upper whisker=uw,
    lower quartile=lq,
    upper quartile=uq,
    median=med}, boxplot prepared] 
    coordinates {}; 
        
	\addplot+[boxplot prepared from table={table=\msespnntrain,row=3,
    lower whisker=lw,
    upper whisker=uw,
    lower quartile=lq,
    upper quartile=uq,
    median=med}, boxplot prepared] 
    coordinates {}; 
	\addplot+[boxplot prepared from table={table=\msespnntest,row=3,
    lower whisker=lw,
    upper whisker=uw,
    lower quartile=lq,
    upper quartile=uq,
    median=med}, boxplot prepared] 
    coordinates {};
    
  \end{semilogyaxis}
  
\end{tikzpicture}
\caption{Box plots for the relative L2 error for all the rollout snapshots of the Couette flow in both train and test cases. The state variables represented are (a) position, (b) velocity, (c) energy and (d) conformation tensor.}
\label{fig:couette_box}
\end{figure*}

The rollout results for the Couette flow are presented in \myfigref{fig:couette_box}. A substantial improvement is shown in the present approach over the two other methods, which remain in a similar performance. Note that the skewed distributions towards higher errors on each box is due to the error accumulation on snapshots further in time from the starting conditions, where errors are lower. \myfigref{fig:cons} (left) shows that the degeneracy conditions imposed by our method ensure the thermodynamical consistency of the learnt energy and entropy potentials.

\subsection{Viscoelastic bending beam}\label{sec:beam}

\subsubsection{Description}

The next example is a viscoelastic cantilever beam subjected by a bending force. The material is characterized by a single-term polynomial strain energy potential, described by the following equation
\begin{equation}
U=C_{10}(\overline{I}_1-3)+C_{01}(\overline{I}_2-3)+\frac{1}{D_1}(J_{el}-1)^2
\end{equation}
where $U$ is the strain energy potential, $J_{el}$ is the elastic volume ratio, $\overline{I}_1$ and $\overline{I}_2$ are the two invariants of the left Cauchy-Green deformation tensor, $C_{10}$ and $C_{01}$ are shear material constants and $D_1$ is the material compressibility parameter. The viscoelastic component is described by a two-term Prony series of the dimensionless shear relaxation modulus,
\begin{equation}
g_R(t)=1-\bar{g}_1(1-e^{\frac{-t}{\tau_1}})-\bar{g}_2(1-e^{\frac{-t}{\tau_2}}),
\end{equation}
with relaxation coefficients of $\bar{g}_1$ and $\bar{g}_2$, and relaxation times of $\tau_1$ and $\tau_2$.

\begin{figure}[h]
\centering
\includegraphics[width=0.45\textwidth]{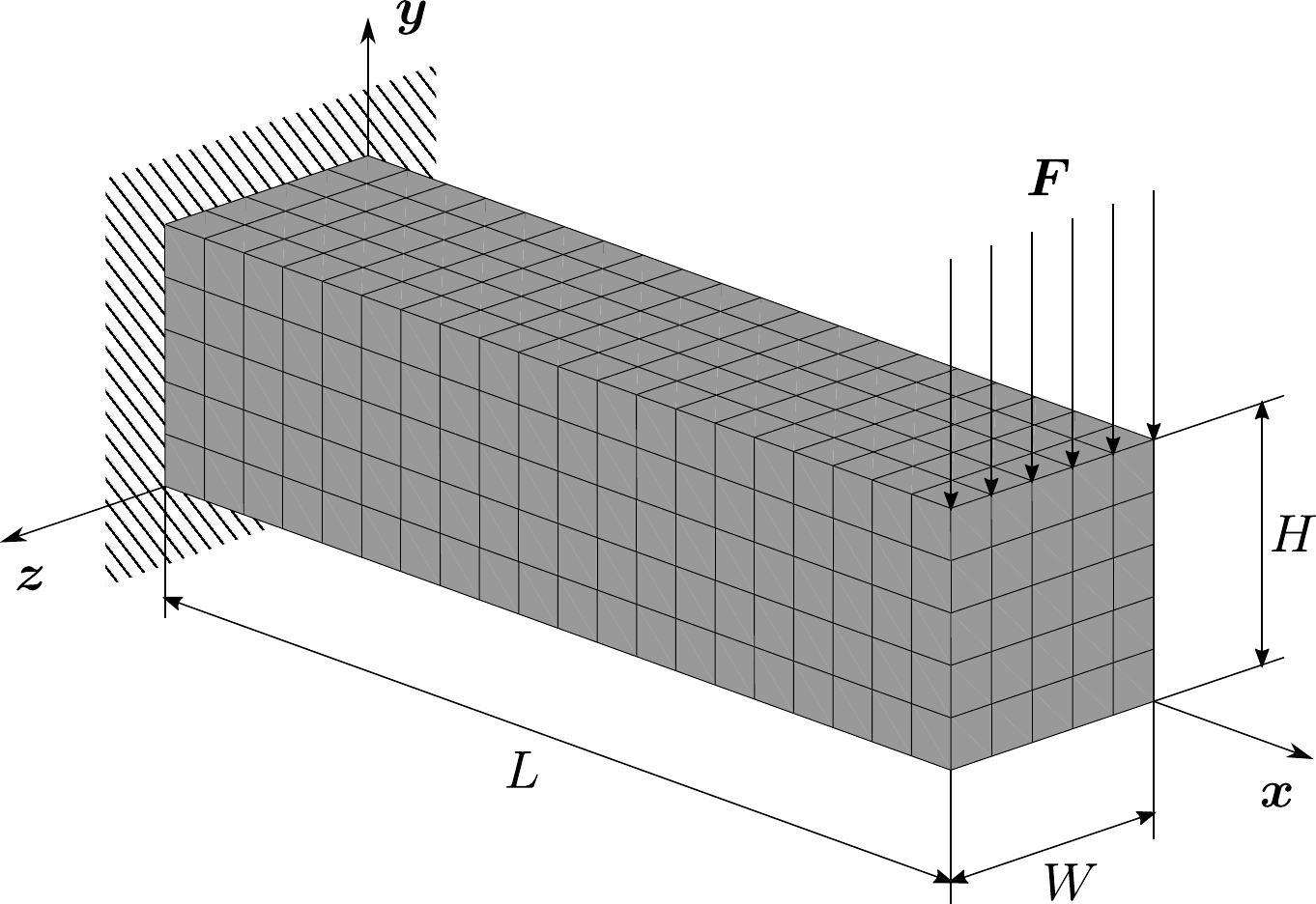} 
\caption{Viscoelastic beam problem with a load case. The load position and direction are modified on each simulation, obtaining different stress fields.\label{fig:beam}}
\end{figure}

The state variables for the viscoelastic beam on each node are the position $\bs{q}$, velocity $\bs{v}$ and stress tensor $\bs{\sigma}$,
\begin{equation}\label{eq:beam_z}
\mathcal{S}=\{\bs{z}=(\bs{q},\bs{v},\bs{\sigma})\in\mathbb{R}^3\times\mathbb{R}^3\times\mathbb{R}^6\}.
\end{equation}
The relative deformed position is included into the edge feature vector whereas the rest of the variables are part of the node feature vector. An additional one-hot vector $\bs{n}$ is added to the node features in order to represent the encastre and beam nodes. The external load vector $\bs{F}$ is included in the node processor MLP as an external interaction. No global feature vector is needed in this case, resulting in the following feature vectors:
\begin{equation}\label{eq:beam_feat}
\bs{e}_{ij}=(\bs{q}_i-\bs{q}_j,\Vert\bs{q}_{i}-\bs{q}_{j}\Vert_2),\quad\bs{v}_i=(\bs{v},\bs{\sigma},\bs{n}).
\end{equation}

\subsubsection{Database and Hyperparameters}

The prismatic beam dimensions are $H=10$, $W=10$ and $L=40$, discretized in $N_e=500$ hexahedral linear brick elements and $N=756$ nodes. The material hyperelastic and viscoelastic parameters are $C_{10}=1.5\e{5}$, $C_{01}=5\e{3}$, $D_1=10^{-7}$ and $\bar{g}_1=0.3$, $\bar{g}_2=0.49$, $\tau_1=0.2$, $\tau_2=0.5$ respectively. A distributed load of $F=10^5$ is applied in $N_\text{sim}=52$ different positions with an orientation perpendicular to the solid surface. The quasi-static simulation is discretized in $N_T=20$ time increments of $\Delta t=5\e{-2}$. 

Following \myeqref{eq:beam_feat}, the dimensions of the graph feature vectors are $F_e=4$, $F_v=11$ and $F_g=0$. The hidden dimension of the node and edge latent vectors is $F_h=50$. The learning rate is set to $l_r=10^{-4}$ with decreasing order of magnitude on epochs $600$ and $1200$, and a total number of $N_\text{epoch}=1800$. The training noise variance is set to $\sigma_\text{noise}^2=10^{-5}$.

\subsubsection{Results}

\begin{figure*}[h]
\centering
\includegraphics[width=\textwidth]{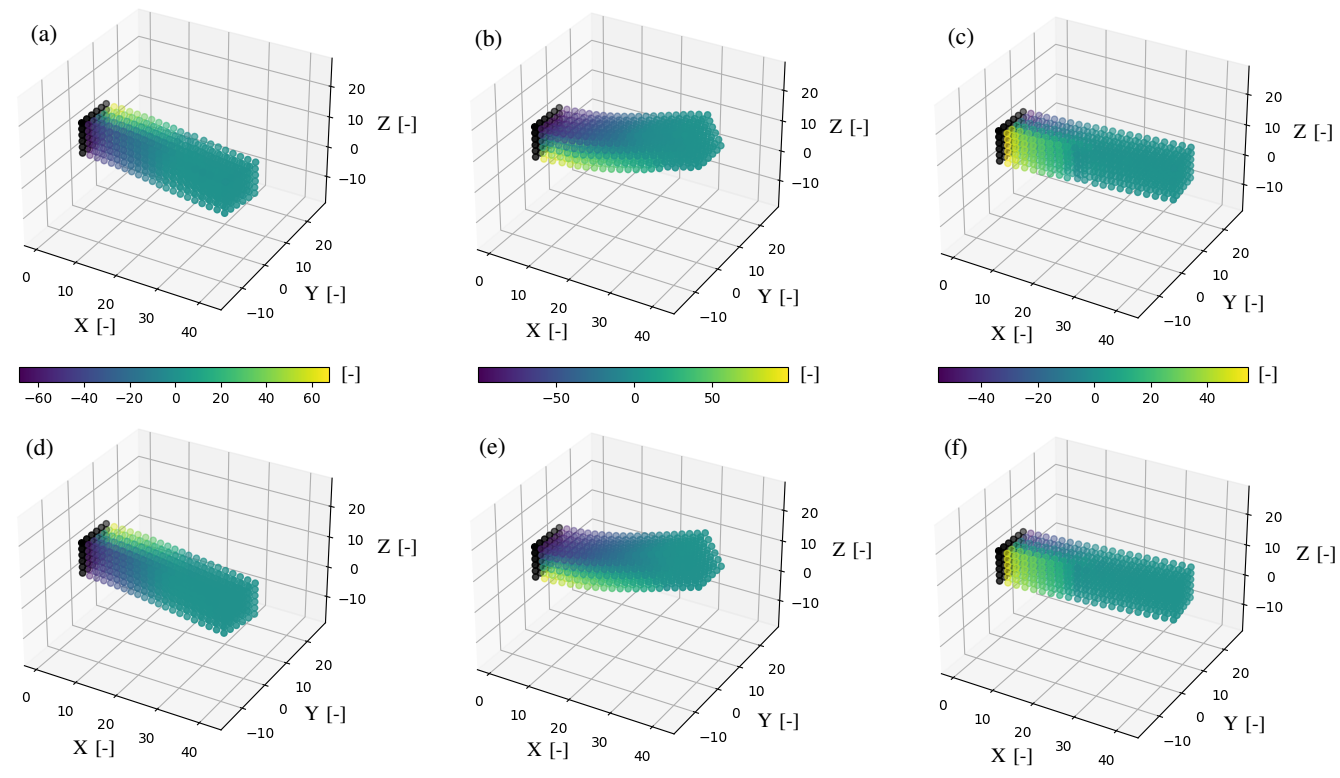}
\caption{(a), (b) and (c): Representation of a snapshot of three test simulations, i.e. not seen by the network on training, of the bending beam problem. (c), (d) and (e): Their respective ground truth simulations. The color code represents the $xx$ component of the dimensionless stress tensor, scaled $\times 0.001$.}
\label{fig:beam_results}
\end{figure*}

The rollout results for the bending viscoelastic beam are presented in \myfigref{fig:beam_box}. The errors achieved by the present approach are again below the other two methods. The beam deformed configuration of three different test simulation snapshots are represented in \myfigref{fig:beam_results}, with the color code representing the $xx$ component of the stress tensor. Similarly to the previous case, \myfigref{fig:cons} (center) shows the thermodynamical consistency of our dynamical integration.

\subsection{Flow past a cylinder}\label{sec:cyl}

\subsubsection{Description}

The last example consists of a viscous unsteady flow past a cylinder obstacle. The flow conditions are set to obtain varying Reynolds regimes, which result in Kármán vortex street and therefore a periodic behaviour in the steady state.  

\begin{figure}[h]
\centerline{\includegraphics[width=0.5\textwidth]{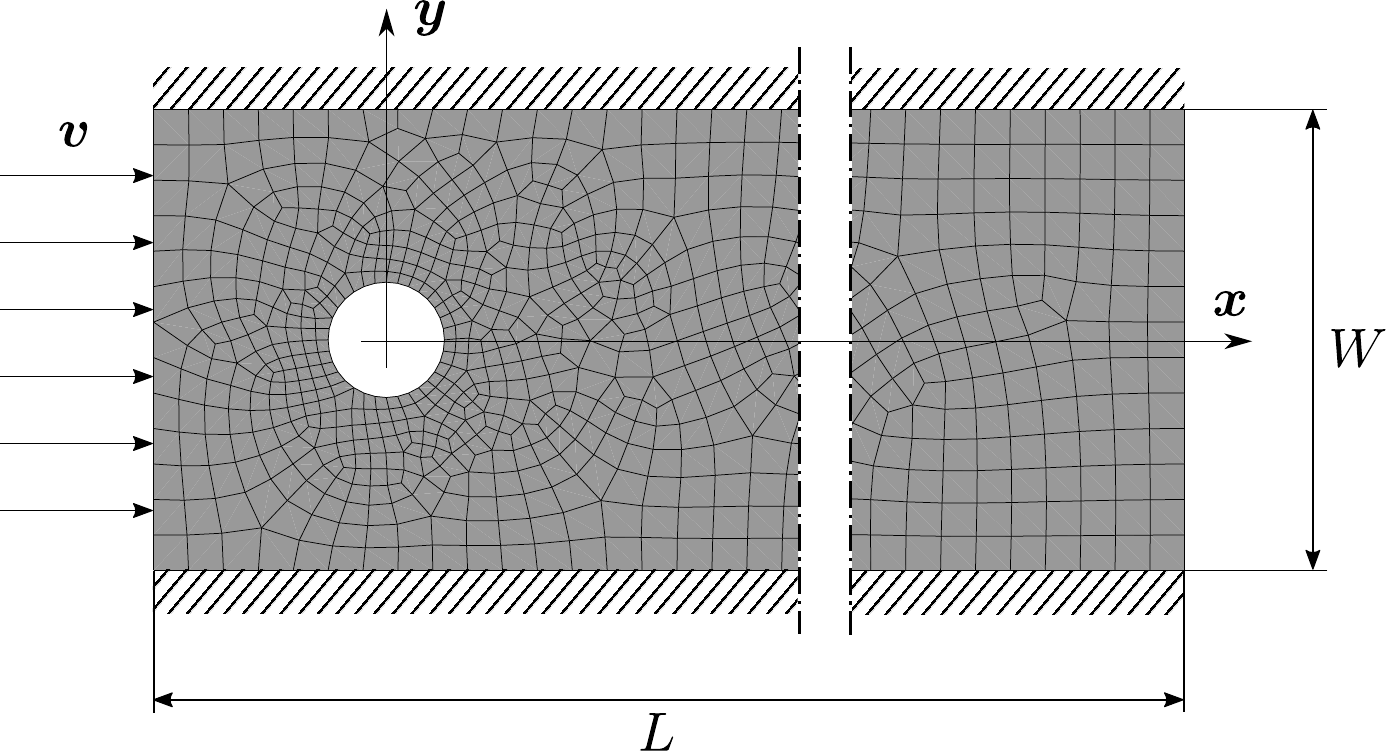}}
\caption{Unsteady flow past a cylinder obstacle. The flow velocity and cylinder obstacle position are varied to obtain different Reynolds numbers and flow profiles.\label{fig:cyl}}
\end{figure}

The state variables for the flow past a cylinder are the velocity $\bs{v}$ and the pressure field $P$,
\begin{equation}\label{eq:cyl_z}
\mathcal{S}=\{\bs{z}=(\bs{v},P)\in\mathbb{R}^2\times\mathbb{R}\}.
\end{equation}
The flow is computed with an Eulerian description of the output fields. Thus, the nodal coordinates ($\bs{q}^0$) are fixed in space and considered as edge features, whereas the whole state variables are assigned to the node features. An additional one-hot vector $\bs{n}$ is added to the node features in order to represent the inlet/outlet, walls or fluid nodes. No global feature vector is needed in this case, resulting in the following feature vectors:
\begin{equation}\label{eq:cyl_feat}
\bs{e}_{ij}=(\bs{q}^0_i-\bs{q}^0_j,\Vert\bs{q}^0_{i}-\bs{q}^0_{j}\Vert_2),\quad\bs{v}_i=(\bs{v},P,\bs{n}).
\end{equation}

\subsubsection{Database and Hyperparameters}

The ground truth simulations are computed solving the 2D Navier Stokes equations. Six different obstacle positions are simulated with varying fluid discretization, which consist of approximately $N_e=1100$ quadrilateral elements and $N=1200$ nodes. No-slip conditions are forced in the stream walls and the cylinder obstacle. The fluid has a density of $\rho=1$ and a dynamic viscosity of $\mu=10^{-3}$. The variable freestream velocity is contained within the interval $\bs{v}\in[1,2]$, summing a total of $N_\text{sim}=30$ cases. The unsteady simulation is discretized in $N_T=300$ time increments of $\Delta t=10^{-2}$.

Following \myeqref{eq:cyl_feat}, the dimensions of the graph feature vectors are $F_e=3$, $F_v=8$ and $F_g=0$. The hidden dimension of the node and edge latent vectors is $F_h=128$. The learning rate is set to $l_r=10^{-4}$ with decreasing order of magnitude on epochs $600$ and $1200$, and a total number of $N_\text{epoch}=2000$. The training noise variance is set to $\sigma_\text{noise}^2=4\e{-4}$.

\subsubsection{Results}

\begin{figure*}[h]
\centering
\includegraphics[width=\textwidth]{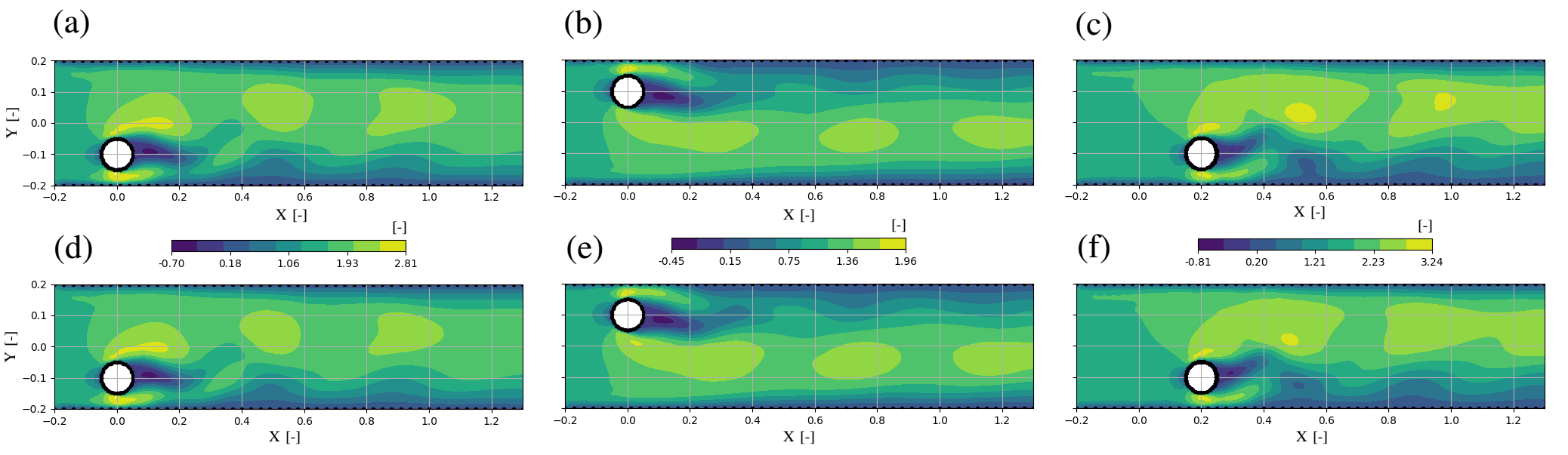}
\caption{(a), (b) and (c): Representation of a snapshot of three test simulations, i.e. not seen by the network on training, of the cylinder flow problem. (c), (d) and (e): Their respective ground truth simulations. The color code represents the $x$ component of the dimensionless velocity field.}
\label{fig:cyl_results}
\end{figure*}

\begin{figure*}[h]
\centering
\begin{tikzpicture}
\pgfplotsset{width=0.3\textwidth, height=6cm}
\begin{axis}[name=plot1,
  grid=major, 
  grid style={dashed,gray!30}, 
  xlabel={(a) Couette flow},
  ylabel={$E,S$ [-]},
  ticklabel style={font=\scriptsize}]
  
  \addplot+[mark repeat = 12] table [y=E, x=snap]{results/conservation/cons_couette.txt};
  \addplot+[mark repeat = 12] table [y=S, x=snap]{results/conservation/cons_couette.txt};
	
\end{axis}
\end{tikzpicture}
~
\begin{tikzpicture}
\pgfplotsset{width=0.3\textwidth, height=6cm}
\begin{axis}[name=plot2,
  grid=major, 
  grid style={dashed,gray!30}, 
  xlabel={(b) Bending beam},
  ticklabel style={font=\scriptsize},
  ]
  
  \addplot+[mark repeat = 2] table [y=E, x=snap]{results/conservation/cons_beam.txt};
  \addplot+[mark repeat = 2] table [y=S, x=snap]{results/conservation/cons_beam.txt};
	
\end{axis}
\end{tikzpicture}
~
\begin{tikzpicture}
\pgfplotsset{width=0.3\textwidth, height=6cm}
\begin{axis}[name=plot3,
  grid=major, 
  grid style={dashed,gray!30}, 
  xlabel={(c) Cylinder flow},
  ticklabel style={font=\scriptsize},
  legend pos=outer north east,
  legend cell align=left,
  ]
  
  \addplot+[mark repeat = 24] table [y=E, x=snap]{results/conservation/cons_cylinder.txt};
  \addplot+[mark repeat = 24] table [y=S, x=snap]{results/conservation/cons_cylinder.txt};
	\legend{Energy, Entropy}
\end{axis}

\end{tikzpicture}
\caption{Conservation of energy and non-decreasing entropy potentials for a test case of the (a) Couette flow, (b) bending beam and (c) cylinder flow. Both quantities are averaged across all graph nodes for visualization.}
\label{fig:cons}
\end{figure*}
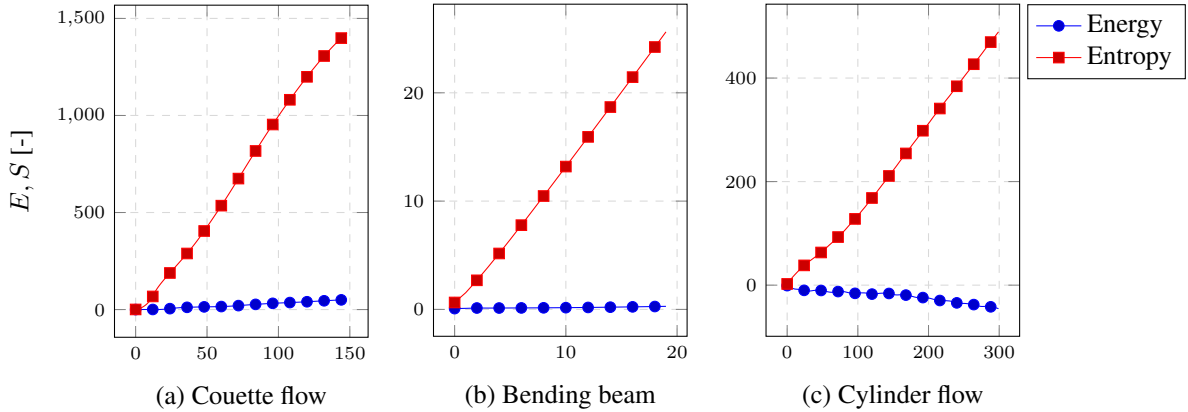

\begin{figure}[h]
\centering
\pgfplotstableread{results/cylinder_box/mse_ours_train.txt}\mseourstrain
\pgfplotstableread{results/cylinder_box/mse_ours_test.txt}\mseourstest
\pgfplotstableread{results/cylinder_box/mse_gnn_train.txt}\msegnntrain
\pgfplotstableread{results/cylinder_box/mse_gnn_test.txt}\msegnntest
\pgfplotstableread{results/cylinder_box/mse_spnn_train.txt}\msespnntrain
\pgfplotstableread{results/cylinder_box/mse_spnn_test.txt}\msespnntest

\begin{tikzpicture}
\pgfplotsset{width=\textwidth, height=6cm}

  \begin{semilogyaxis}
  [xshift=0cm,xlabel=(a) Velocity,
  area legend, boxplot/draw direction=y,
  grid=major, 
  grid style={dashed,gray!30}, 
  cycle list={{red},{blue}},
  boxplot={draw position={1/3 + floor(\plotnumofactualtype/2) + 1/3*mod(\plotnumofactualtype,2)},box extend=0.3,},
  x=1cm,xtick={0,1,2,...,10},x tick label as interval,
  xticklabels={{Ours},{GNN},{SPNN}},
  ticklabel style={font=\scriptsize},
  ylabel={Relative L2 error},
  ] 
	\addplot+[boxplot prepared from table={table=\mseourstrain,row=0,
    lower whisker=lw,
    upper whisker=uw,
    lower quartile=lq,
    upper quartile=uq,
    median=med}, boxplot prepared]
    coordinates {};
	\addplot+[boxplot prepared from table={table=\mseourstest,row=0,
    lower whisker=lw,
    upper whisker=uw,
    lower quartile=lq,
    upper quartile=uq,
    median=med}, boxplot prepared] 
    coordinates {}; 
    
	\addplot+[boxplot prepared from table={table=\msegnntrain,row=0,
    lower whisker=lw,
    upper whisker=uw,
    lower quartile=lq,
    upper quartile=uq,
    median=med}, boxplot prepared] 
    coordinates {}; 
	\addplot+[boxplot prepared from table={table=\msegnntest,row=0,
    lower whisker=lw,
    upper whisker=uw,
    lower quartile=lq,
    upper quartile=uq,
    median=med}, boxplot prepared] 
    coordinates {}; 
        
	\addplot+[boxplot prepared from table={table=\msespnntrain,row=0,
    lower whisker=lw,
    upper whisker=uw,
    lower quartile=lq,
    upper quartile=uq,
    median=med}, boxplot prepared] 
    coordinates {}; 
	\addplot+[boxplot prepared from table={table=\msespnntest,row=0,
    lower whisker=lw,
    upper whisker=uw,
    lower quartile=lq,
    upper quartile=uq,
    median=med}, boxplot prepared] 
    coordinates {};
    
  \end{semilogyaxis}
  
  \begin{semilogyaxis}
  [xshift=4cm,xlabel=(b) Pressure,
  area legend, boxplot/draw direction=y,
  grid=major, 
  grid style={dashed,gray!30}, 
  cycle list={{red},{blue}},
  boxplot={draw position={1/3 + floor(\plotnumofactualtype/2) + 1/3*mod(\plotnumofactualtype,2)},box extend=0.3,},
  x=1cm,xtick={0,1,2,...,10},x tick label as interval,
  xticklabels={{Ours},{GNN},{SPNN}},
  ticklabel style={font=\scriptsize},
  custom legend,legend pos=north west,legend cell align=left,
  legend entries = {Train, Test},
  ] 
	\addplot+[boxplot prepared from table={table=\mseourstrain,row=1,
    lower whisker=lw,
    upper whisker=uw,
    lower quartile=lq,
    upper quartile=uq,
    median=med}, boxplot prepared] 
    coordinates {};
	\addplot+[boxplot prepared from table={table=\mseourstest,row=1,
    lower whisker=lw,
    upper whisker=uw,
    lower quartile=lq,
    upper quartile=uq,
    median=med}, boxplot prepared] 
    coordinates {}; 
    
	\addplot+[boxplot prepared from table={table=\msegnntrain,row=1,
    lower whisker=lw,
    upper whisker=uw,
    lower quartile=lq,
    upper quartile=uq,
    median=med}, boxplot prepared] 
    coordinates {}; 
	\addplot+[boxplot prepared from table={table=\msegnntest,row=1,
    lower whisker=lw,
    upper whisker=uw,
    lower quartile=lq,
    upper quartile=uq,
    median=med}, boxplot prepared] 
    coordinates {}; 
        
	\addplot+[boxplot prepared from table={table=\msespnntrain,row=1,
    lower whisker=lw,
    upper whisker=uw,
    lower quartile=lq,
    upper quartile=uq,
    median=med}, boxplot prepared] 
    coordinates {}; 
	\addplot+[boxplot prepared from table={table=\msespnntest,row=1,
    lower whisker=lw,
    upper whisker=uw,
    lower quartile=lq,
    upper quartile=uq,
    median=med}, boxplot prepared] 
    coordinates {};
    
  \end{semilogyaxis}

\end{tikzpicture}
\caption{Box plots for the relative L2 error for all the rollout snapshots of the cylinder flow in both train and test cases. The state variables represented are (a) velocity and (b) pressure.}
\label{fig:cyl_box}
\end{figure}
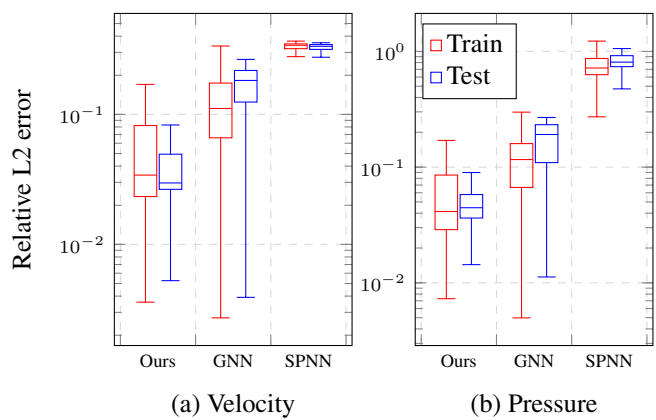

The rollout results for the flow past a cylinder problem are presented in \myfigref{fig:cyl_box}. In this example the domain varies significantly, using a different unstructured mesh for each simulation. Thus, the graph-based architectures outperform the vanilla SPNN, which is meant for fixed structured problems. Considering the other two methods, our approach outperforms the standard GNN architecture due to the metriplectic structure imposition over the dynamical problem, as depicted in \myfigref{fig:cons}. A single snapshot of the whole rollout of three different test simulations are represented in \myfigref{fig:cyl_results}, with the color code representing the $x$ component of the velocity field.

\section{Conclusions}\label{sec:conc}

We have presented a method to predict the time evolution of an arbitrary dynamical system based on two inductive biases. The metriplectic bias ensures the correct thermodynamic structure of the integrator based on the GENERIC formalism, whose operators and potentials are estimated using computations over graphs, i.e. exploiting the geometric structure of the problem. The results show relative mean errors of less than 3\% in all the tested examples, outperforming two other state-of-the-art techniques based on only physics-informed and geometric deep learning respectively. These results confirm that both biases are necessary to achieve higher precision in the predicted simulations. The use of both techniques combine the computational power of geometric deep learning with the rigorous foundation of the GENERIC formalism, which ensure the thermodynamical consistency of the results.

The limitations of the presented technique are related to the computational complexity of the model. Large simulations with fine grids require a high amount of message passing to get the information across the whole domain, or a very fine time discretization, which both result in a high computational cost. Similarly, high-speed phenomena in relation to the wave velocity of the medium might be impossible to model.

Future work may overcome the stated limitations by combining graph representations with model order reduction techniques, such as autoencoders or U-net architectures \cite{gao2019graph,yu2019st}. The idea is to replace deep message passing with various coarse-graining steps, allowing the boundary information to reach every node in the simulation domain while reducing the number of parameters of the neural network. Another interesting topic to extend our work is to improve the generalization and decrease the amount of training data via equivariant arquitectures \cite{keriven2019universal,satorras2021n}, which avoid data augmentation by exploiting the invariance to certain groups such as rotations $SO(3)$ or general Euclidean transformations $E(3)$. As the present work is only limited to in-silico experiments, future work may extend the proposed method to measured datasets in real-world applications, such as digital twins of industrial processes or real-time augmented/virtual reality environments.

\begin{figure*}[h]
\centering
\pgfplotstableread{results/beam_box/mse_ours_train.txt}\mseourstrain
\pgfplotstableread{results/beam_box/mse_ours_test.txt}\mseourstest
\pgfplotstableread{results/beam_box/mse_gnn_train.txt}\msegnntrain
\pgfplotstableread{results/beam_box/mse_gnn_test.txt}\msegnntest
\pgfplotstableread{results/beam_box/mse_spnn_train.txt}\msespnntrain
\pgfplotstableread{results/beam_box/mse_spnn_test.txt}\msespnntest

\begin{tikzpicture}
\pgfplotsset{width=\textwidth, height=6cm}

  \begin{semilogyaxis}
  [xshift=0cm,xlabel=(a) Position,
  area legend, boxplot/draw direction=y,
  grid=major, 
  grid style={dashed,gray!30}, 
  cycle list={{red},{blue}},
  boxplot={draw position={1/3 + floor(\plotnumofactualtype/2) + 1/3*mod(\plotnumofactualtype,2)},box extend=0.3,},
  x=1cm,xtick={0,1,2,...,10},x tick label as interval,
  xticklabels={{Ours},{GNN},{SPNN}},
  ticklabel style={font=\scriptsize},
  ymax=0.1,
  ylabel={Relative L2 error},
  ] 
	\addplot+[boxplot prepared from table={table=\mseourstrain,row=0,
    lower whisker=lw,
    upper whisker=uw,
    lower quartile=lq,
    upper quartile=uq,
    median=med}, boxplot prepared]
    coordinates {};
	\addplot+[boxplot prepared from table={table=\mseourstest,row=0,
    lower whisker=lw,
    upper whisker=uw,
    lower quartile=lq,
    upper quartile=uq,
    median=med}, boxplot prepared] 
    coordinates {}; 
    
	\addplot+[boxplot prepared from table={table=\msegnntrain,row=0,
    lower whisker=lw,
    upper whisker=uw,
    lower quartile=lq,
    upper quartile=uq,
    median=med}, boxplot prepared] 
    coordinates {}; 
	\addplot+[boxplot prepared from table={table=\msegnntest,row=0,
    lower whisker=lw,
    upper whisker=uw,
    lower quartile=lq,
    upper quartile=uq,
    median=med}, boxplot prepared] 
    coordinates {}; 
        
	\addplot+[boxplot prepared from table={table=\msespnntrain,row=0,
    lower whisker=lw,
    upper whisker=uw,
    lower quartile=lq,
    upper quartile=uq,
    median=med}, boxplot prepared] 
    coordinates {}; 
	\addplot+[boxplot prepared from table={table=\msespnntest,row=0,
    lower whisker=lw,
    upper whisker=uw,
    lower quartile=lq,
    upper quartile=uq,
    median=med}, boxplot prepared] 
    coordinates {};
    
  \end{semilogyaxis}
  
  \begin{semilogyaxis}
  [xshift=4cm,xlabel=(b) Velocity,
  area legend, boxplot/draw direction=y,
  grid=major, 
  grid style={dashed,gray!30}, 
  cycle list={{red},{blue}},
  boxplot={draw position={1/3 + floor(\plotnumofactualtype/2) + 1/3*mod(\plotnumofactualtype,2)},box extend=0.3,},
  x=1cm,xtick={0,1,2,...,10},x tick label as interval,
  xticklabels={{Ours},{GNN},{SPNN}},
  ticklabel style={font=\scriptsize},
  ymin=0.001,ymax=1,
  ] 
	\addplot+[boxplot prepared from table={table=\mseourstrain,row=1,
    lower whisker=lw,
    upper whisker=uw,
    lower quartile=lq,
    upper quartile=uq,
    median=med}, boxplot prepared] 
    coordinates {};
	\addplot+[boxplot prepared from table={table=\mseourstest,row=1,
    lower whisker=lw,
    upper whisker=uw,
    lower quartile=lq,
    upper quartile=uq,
    median=med}, boxplot prepared] 
    coordinates {}; 
    
	\addplot+[boxplot prepared from table={table=\msegnntrain,row=1,
    lower whisker=lw,
    upper whisker=uw,
    lower quartile=lq,
    upper quartile=uq,
    median=med}, boxplot prepared] 
    coordinates {}; 
	\addplot+[boxplot prepared from table={table=\msegnntest,row=1,
    lower whisker=lw,
    upper whisker=uw,
    lower quartile=lq,
    upper quartile=uq,
    median=med}, boxplot prepared] 
    coordinates {}; 
        
	\addplot+[boxplot prepared from table={table=\msespnntrain,row=1,
    lower whisker=lw,
    upper whisker=uw,
    lower quartile=lq,
    upper quartile=uq,
    median=med}, boxplot prepared] 
    coordinates {}; 
	\addplot+[boxplot prepared from table={table=\msespnntest,row=1,
    lower whisker=lw,
    upper whisker=uw,
    lower quartile=lq,
    upper quartile=uq,
    median=med}, boxplot prepared] 
    coordinates {};
    
  \end{semilogyaxis}

  \begin{semilogyaxis}
  [xshift=8cm,xlabel=(c) Stress Tensor,
  area legend, boxplot/draw direction=y,
  grid=major, 
  grid style={dashed,gray!30}, 
  cycle list={{red},{blue}},
  boxplot={draw position={1/3 + floor(\plotnumofactualtype/2) + 1/3*mod(\plotnumofactualtype,2)},box extend=0.3,},
  x=1cm,xtick={0,1,2,...,10},x tick label as interval,
  xticklabels={{Ours},{GNN},{SPNN}},
  ticklabel style={font=\scriptsize},
  ymin=0.01,ymax=1,
  custom legend,legend pos=outer north east,legend cell align=left,
  legend entries = {Train, Test},
  ] 
	\addplot+[boxplot prepared from table={table=\mseourstrain,row=2,
    lower whisker=lw,
    upper whisker=uw,
    lower quartile=lq,
    upper quartile=uq,
    median=med}, boxplot prepared] 
    coordinates {};
	\addplot+[boxplot prepared from table={table=\mseourstest,row=2,
    lower whisker=lw,
    upper whisker=uw,
    lower quartile=lq,
    upper quartile=uq,
    median=med}, boxplot prepared] 
    coordinates {}; 
    
	\addplot+[boxplot prepared from table={table=\msegnntrain,row=2,
    lower whisker=lw,
    upper whisker=uw,
    lower quartile=lq,
    upper quartile=uq,
    median=med}, boxplot prepared] 
    coordinates {}; 
	\addplot+[boxplot prepared from table={table=\msegnntest,row=2,
    lower whisker=lw,
    upper whisker=uw,
    lower quartile=lq,
    upper quartile=uq,
    median=med}, boxplot prepared] 
    coordinates {}; 
        
	\addplot+[boxplot prepared from table={table=\msespnntrain,row=2,
    lower whisker=lw,
    upper whisker=uw,
    lower quartile=lq,
    upper quartile=uq,
    median=med}, boxplot prepared] 
    coordinates {}; 
	\addplot+[boxplot prepared from table={table=\msespnntest,row=2,
    lower whisker=lw,
    upper whisker=uw,
    lower quartile=lq,
    upper quartile=uq,
    median=med}, boxplot prepared] 
    coordinates {};
    
  \end{semilogyaxis}

\end{tikzpicture}
\caption{Box plots for the relative L2 error for all the rollout snapshots of the bending viscoelastic beam in both train and test cases. The state variables represented are (a) position, (b) velocity, and (c) stress tensor.}
\label{fig:beam_box}
\end{figure*}
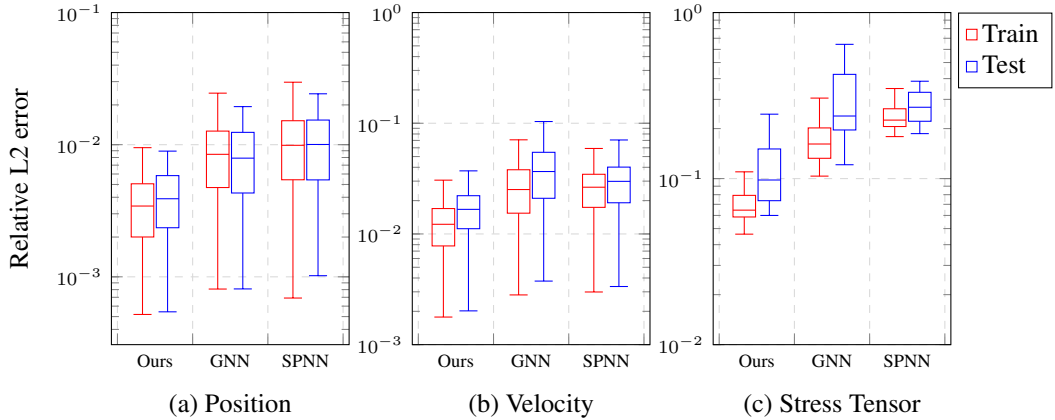

\bibliography{GRAPH-GENERIC-ARXIV}

\begin{thebibliography}{10}

\bibitem{raissi2019physics}
Maziar Raissi, Paris Perdikaris, and George~E Karniadakis.
\newblock Physics-informed neural networks: A deep learning framework for
  solving forward and inverse problems involving nonlinear partial differential
  equations.
\newblock {\em Journal of Computational Physics}, 378:686--707, 2019.

\bibitem{cranmer2020discovering}
Miles Cranmer, Alvaro Sanchez~Gonzalez, Peter Battaglia, Rui Xu, Kyle Cranmer,
  David Spergel, and Shirley Ho.
\newblock Discovering symbolic models from deep learning with inductive biases.
\newblock {\em Advances in Neural Information Processing Systems},
  33:17429--17442, 2020.

\bibitem{moya2021physics}
Beatriz Moya, Alberto Badias, David Gonzalez, Francisco Chinesta, and Elias
  Cueto.
\newblock Physics perception in sloshing scenes with guaranteed thermodynamic
  consistency.
\newblock {\em arXiv preprint arXiv:2106.13301}, 2021.

\bibitem{lecun1999object}
Yann LeCun, Patrick Haffner, L{\'e}on Bottou, and Yoshua Bengio.
\newblock Object recognition with gradient-based learning.
\newblock In {\em Shape, contour and grouping in computer vision}, pages
  319--345. Springer, 1999.

\bibitem{rumelhart1985learning}
David~E Rumelhart, Geoffrey~E Hinton, and Ronald~J Williams.
\newblock Learning internal representations by error propagation.
\newblock Technical report, California Univ San Diego La Jolla Inst for
  Cognitive Science, 1985.

\bibitem{bronstein2017geometric}
Michael~M Bronstein, Joan Bruna, Yann LeCun, Arthur Szlam, and Pierre
  Vandergheynst.
\newblock Geometric deep learning: going beyond euclidean data.
\newblock {\em IEEE Signal Processing Magazine}, 34(4):18--42, 2017.

\bibitem{raissi2018hidden}
Maziar Raissi and George~Em Karniadakis.
\newblock Hidden physics models: Machine learning of nonlinear partial
  differential equations.
\newblock {\em Journal of Computational Physics}, 357:125--141, 2018.

\bibitem{greydanus2019hamiltonian}
Samuel~J Greydanus, Misko Dzumba, and Jason Yosinski.
\newblock Hamiltonian neural networks.
\newblock 2019.

\bibitem{sanchez2019hamiltonian}
Alvaro Sanchez-Gonzalez, Victor Bapst, Kyle Cranmer, and Peter Battaglia.
\newblock Hamiltonian graph networks with ode integrators.
\newblock {\em arXiv preprint arXiv:1909.12790}, 2019.

\bibitem{mattheakis2020hamiltonian}
Marios Mattheakis, David Sondak, Akshunna~S Dogra, and Pavlos Protopapas.
\newblock Hamiltonian neural networks for solving differential equations.
\newblock {\em arXiv preprint arXiv:2001.11107}, 2020.

\bibitem{desai2021port}
Shaan Desai, Marios Mattheakis, David Sondak, Pavlos Protopapas, and Stephen
  Roberts.
\newblock Port-hamiltonian neural networks for learning explicit time-dependent
  dynamical systems.
\newblock {\em arXiv preprint arXiv:2107.08024}, 2021.

\bibitem{chen2019symplectic}
Zhengdao Chen, Jianyu Zhang, Martin Arjovsky, and L{\'e}on Bottou.
\newblock Symplectic recurrent neural networks.
\newblock {\em arXiv preprint arXiv:1909.13334}, 2019.

\bibitem{ottinger1997dynamics}
Hans~Christian {\"O}ttinger and Miroslav Grmela.
\newblock Dynamics and thermodynamics of complex fluids. ii. illustrations of a
  general formalism.
\newblock {\em Physical Review E}, 56(6):6633, 1997.

\bibitem{grmela1997dynamics}
Miroslav Grmela and Hans~Christian {\"O}ttinger.
\newblock Dynamics and thermodynamics of complex fluids. i. development of a
  general formalism.
\newblock {\em Physical Review E}, 56(6):6620, 1997.

\bibitem{hernandez2021structure}
Quercus Hern{\'a}ndez, Alberto Bad{\'\i}as, David Gonz{\'a}lez, Francisco
  Chinesta, and El{\'\i}as Cueto.
\newblock Structure-preserving neural networks.
\newblock {\em Journal of Computational Physics}, 426:109950, 2021.

\bibitem{hernandez2021deep}
Quercus Hernandez, Alberto Badias, David Gonzalez, Francisco Chinesta, and
  Elias Cueto.
\newblock Deep learning of thermodynamics-aware reduced-order models from data.
\newblock {\em Computer Methods in Applied Mechanics and Engineering},
  379:113763, 2021.

\bibitem{lee2021machine}
Kookjin Lee, Nathaniel~A Trask, and Panos Stinis.
\newblock Machine learning structure preserving brackets for forecasting
  irreversible processes.
\newblock {\em arXiv preprint arXiv:2106.12619}, 2021.

\bibitem{zhang2021gfinns}
Zhen Zhang, Yeonjong Shin, and George~Em Karniadakis.
\newblock Gfinns: Generic formalism informed neural networks for deterministic
  and stochastic dynamical systems.
\newblock {\em arXiv preprint arXiv:2109.00092}, 2021.

\bibitem{hu2021forcenet}
Weihua Hu, Muhammed Shuaibi, Abhishek Das, Siddharth Goyal, Anuroop Sriram,
  Jure Leskovec, Devi Parikh, and C~Lawrence Zitnick.
\newblock Forcenet: A graph neural network for large-scale quantum
  calculations.
\newblock {\em arXiv preprint arXiv:2103.01436}, 2021.

\bibitem{shlomi2020graph}
Jonathan Shlomi, Peter Battaglia, and Jean-Roch Vlimant.
\newblock Graph neural networks in particle physics.
\newblock {\em Machine Learning: Science and Technology}, 2(2):021001, 2020.

\bibitem{ju2020graph}
Xiangyang Ju, Steven Farrell, Paolo Calafiura, Daniel Murnane, Lindsey Gray,
  Thomas Klijnsma, Kevin Pedro, Giuseppe Cerati, Jim Kowalkowski, Gabriel
  Perdue, et~al.
\newblock Graph neural networks for particle reconstruction in high energy
  physics detectors.
\newblock {\em arXiv preprint arXiv:2003.11603}, 2020.

\bibitem{kipf2018neural}
Thomas Kipf, Ethan Fetaya, Kuan-Chieh Wang, Max Welling, and Richard Zemel.
\newblock Neural relational inference for interacting systems.
\newblock In {\em International Conference on Machine Learning}, pages
  2688--2697. PMLR, 2018.

\bibitem{battaglia2018relational}
Peter~W Battaglia, Jessica~B Hamrick, Victor Bapst, Alvaro Sanchez-Gonzalez,
  Vinicius Zambaldi, Mateusz Malinowski, Andrea Tacchetti, David Raposo, Adam
  Santoro, Ryan Faulkner, et~al.
\newblock Relational inductive biases, deep learning, and graph networks.
\newblock {\em arXiv preprint arXiv:1806.01261}, 2018.

\bibitem{sanchez2020learning}
Alvaro Sanchez-Gonzalez, Jonathan Godwin, Tobias Pfaff, Rex Ying, Jure
  Leskovec, and Peter Battaglia.
\newblock Learning to simulate complex physics with graph networks.
\newblock In {\em International Conference on Machine Learning}, pages
  8459--8468. PMLR, 2020.

\bibitem{pfaff2020learning}
Tobias Pfaff, Meire Fortunato, Alvaro Sanchez-Gonzalez, and Peter~W Battaglia.
\newblock Learning mesh-based simulation with graph networks.
\newblock {\em arXiv preprint arXiv:2010.03409}, 2020.

\bibitem{zheng2021deep}
Mianlun Zheng, Yi~Zhou, Duygu Ceylan, and Jernej Barbic.
\newblock A deep emulator for secondary motion of 3d characters.
\newblock In {\em Proceedings of the IEEE/CVF Conference on Computer Vision and
  Pattern Recognition}, pages 5932--5940, 2021.

\bibitem{weinan2017proposal}
E~Weinan.
\newblock A proposal on machine learning via dynamical systems.
\newblock {\em Communications in Mathematics and Statistics}, 5(1):1--11, 2017.

\bibitem{landau1965collected}
Lev~Davidovich Landau.
\newblock {\em Collected papers of LD Landau}.
\newblock Pergamon, 1965.

\bibitem{morrison1986paradigm}
Philip~J Morrison.
\newblock A paradigm for joined hamiltonian and dissipative systems.
\newblock {\em Physica D: Nonlinear Phenomena}, 18(1-3):410--419, 1986.

\bibitem{guha2007metriplectic}
Partha Guha.
\newblock Metriplectic structure, leibniz dynamics and dissipative systems.
\newblock {\em Journal of Mathematical Analysis and Applications},
  326(1):121--136, 2007.

\bibitem{monti2018dual}
Federico Monti, Oleksandr Shchur, Aleksandar Bojchevski, Or~Litany, Stephan
  G{\"u}nnemann, and Michael~M Bronstein.
\newblock Dual-primal graph convolutional networks.
\newblock {\em arXiv preprint arXiv:1806.00770}, 2018.

\bibitem{velivckovic2017graph}
Petar Veli{\v{c}}kovi{\'c}, Guillem Cucurull, Arantxa Casanova, Adriana Romero,
  Pietro Lio, and Yoshua Bengio.
\newblock Graph attention networks.
\newblock {\em arXiv preprint arXiv:1710.10903}, 2017.

\bibitem{zhang2018gaan}
Jiani Zhang, Xingjian Shi, Junyuan Xie, Hao Ma, Irwin King, and Dit-Yan Yeung.
\newblock Gaan: Gated attention networks for learning on large and
  spatiotemporal graphs.
\newblock {\em arXiv preprint arXiv:1803.07294}, 2018.

\bibitem{wang2019dynamic}
Yue Wang, Yongbin Sun, Ziwei Liu, Sanjay~E Sarma, Michael~M Bronstein, and
  Justin~M Solomon.
\newblock Dynamic graph cnn for learning on point clouds.
\newblock {\em Acm Transactions On Graphics (tog)}, 38(5):1--12, 2019.

\bibitem{shi2020point}
Weijing Shi and Raj Rajkumar.
\newblock Point-gnn: Graph neural network for 3d object detection in a point
  cloud.
\newblock In {\em Proceedings of the IEEE/CVF conference on computer vision and
  pattern recognition}, pages 1711--1719, 2020.

\bibitem{gilmer2017neural}
Justin Gilmer, Samuel~S Schoenholz, Patrick~F Riley, Oriol Vinyals, and
  George~E Dahl.
\newblock Neural message passing for quantum chemistry.
\newblock In {\em International Conference on Machine Learning}, pages
  1263--1272. PMLR, 2017.

\bibitem{he2016deep}
Kaiming He, Xiangyu Zhang, Shaoqing Ren, and Jian Sun.
\newblock Deep residual learning for image recognition.
\newblock In {\em Proceedings of the IEEE conference on computer vision and
  pattern recognition}, pages 770--778, 2016.

\bibitem{paszke2017automatic}
Adam Paszke, Sam Gross, Soumith Chintala, Gregory Chanan, Edward Yang, Zachary
  DeVito, Zeming Lin, Alban Desmaison, Luca Antiga, and Adam Lerer.
\newblock Automatic differentiation in pytorch.
\newblock 2017.

\bibitem{teichert2019machine}
Gregory~H Teichert, AR~Natarajan, A~Van~der Ven, and Krishna Garikipati.
\newblock Machine learning materials physics: Integrable deep neural networks
  enable scale bridging by learning free energy functions.
\newblock {\em Computer Methods in Applied Mechanics and Engineering},
  353:201--216, 2019.

\bibitem{hornik1991approximation}
Kurt Hornik.
\newblock Approximation capabilities of multilayer feedforward networks.
\newblock {\em Neural networks}, 4(2):251--257, 1991.

\bibitem{ramachandran2017searching}
Prajit Ramachandran, Barret Zoph, and Quoc~V Le.
\newblock Searching for activation functions.
\newblock {\em arXiv preprint arXiv:1710.05941}, 2017.

\bibitem{misra2019mish}
Diganta Misra.
\newblock Mish: A self regularized non-monotonic neural activation function.
\newblock {\em arXiv preprint arXiv:1908.08681}, 4, 2019.

\bibitem{kingma2014adam}
Diederik~P Kingma and Jimmy Ba.
\newblock Adam: A method for stochastic optimization.
\newblock {\em arXiv preprint arXiv:1412.6980}, 2014.

\bibitem{laso1993calculation}
Manuel Laso and Hans~Christian {\"O}ttinger.
\newblock Calculation of viscoelastic flow using molecular models: the
  connffessit approach.
\newblock {\em Journal of Non-Newtonian Fluid Mechanics}, 47:1--20, 1993.

\bibitem{le2009multiscale}
Claude Le~Bris and Tony Lelievre.
\newblock Multiscale modelling of complex fluids: a mathematical initiation.
\newblock In {\em Multiscale modeling and simulation in science}, pages
  49--137. Springer, 2009.

\bibitem{gao2019graph}
Hongyang Gao and Shuiwang Ji.
\newblock Graph u-nets.
\newblock In {\em international conference on machine learning}, pages
  2083--2092. PMLR, 2019.

\bibitem{yu2019st}
Bing Yu, Haoteng Yin, and Zhanxing Zhu.
\newblock St-unet: A spatio-temporal u-network for graph-structured time series
  modeling.
\newblock {\em arXiv preprint arXiv:1903.05631}, 2019.

\bibitem{keriven2019universal}
Nicolas Keriven and Gabriel Peyr{\'e}.
\newblock Universal invariant and equivariant graph neural networks.
\newblock {\em Advances in Neural Information Processing Systems},
  32:7092--7101, 2019.

\bibitem{satorras2021n}
Victor~Garcia Satorras, Emiel Hoogeboom, and Max Welling.
\newblock E (n) equivariant graph neural networks.
\newblock {\em arXiv preprint arXiv:2102.09844}, 2021.

\end{thebibliography}
\bibliographystyle{unsrt}

\end{document}